%% file: marytts-lrec2018.tex
\pdfoutput=1
\documentclass[10pt, a4paper, final]{article}

\usepackage{lrec}
\usepackage{multicol}

\usepackage[hang,flushmargin]{footmisc}

\usepackage{titlesec}
\renewcommand{\thesection}{\arabic{section}}
\renewcommand{\thesubsection}{\thesection.\arabic{subsection}}

\titlelabel{\thetitle.\quad}

\usepackage{ifxetex}
\ifxetex
  \usepackage{mathspec}
  \setprimaryfont{Times}
  \setallmonofonts{Courier}
\else
  \usepackage[utf8]{inputenc}
\fi
\usepackage{relsize}

\usepackage[english]{babel}
\usepackage{csquotes}

\usepackage{graphicx}
\usepackage{tikz}
\usetikzlibrary{%
  backgrounds,
  bending,
  calc,
  decorations.text,
  fit,
  matrix,
  positioning,
  shadings,
  shapes.multipart,
  shapes.symbols
}
\colorlet{cloudthingiecolor}{blue!10}
\colorlet{projectcolor}{green!20}

\usepackage{subcaption}

\usepackage{tabularx}

\usepackage[inline, shortlabels]{enumitem}

\usepackage[hyphens]{url}
\usepackage[hidelinks, breaklinks, pdfusetitle]{hyperref}
\usepackage[capitalise, nameinlink, noabbrev]{cleveref}

\usepackage[colorinlistoftodos, obeyFinal]{todonotes}
\usepackage[printonlyused, withpage]{acronym}
\newacro{ASR}{automatic speech recognition}
\newacro{BAP}{band aperiodicity}
\newacro{CART}{classification and regression tree}
\newacro{DAW}{digital audio workstation}
\newacro{DFG}{German Research Foundation}
\newacro{DNN}{deep neural network}
\newacro{EST}{Edinburgh Speech Tools}
\newacroindefinite{EST}{\ignorespaces}{the}
\newacro{F0}[F$_0$]{fundamental frequency}
\newacro{FST}{finite state transducer}
\newacro{FOSS}{free, open source software}
\newacro{G2P}{grapheme-to-phoneme}
\newacro{GMM}{Gaussian mixture model}
\newacro{HCI}{human-computer interaction}
\newacro{HMM}{hidden Markov model}
\newacroindefinite{HMM}{an}{a}
\newacro{HTS}{\ac{HMM} based speech synthesis system}
\newacroindefinite{HTS}{\ignorespaces}{the}
\newacro{JRE}{Java Runtime Environment}
\newacro{IPA}{International Phonetic Alphabet}
\newacro{LSTM}{Long Short-Term Memory}
\newacro{MCD}{mel-cepstral distortion}
\newacro{MGC}{mel-generalized cepstrum}
\newacro{MFCC}{mel-frequency cepstral coefficient}
\newacro{MLSA}{mel-generalized log spectrum approximation}
\newacro{MOS}{mean opinion score}
\newacro{MUSHRA}{multiple stimuli with hidden reference and anchor}
\newacro{NLP}{natural language processing}
\newacro{OOV}{out-of-vocabulary}
\newacro{POS}{part-of-speech}
\newacro{RMSE}{root mean square error}
\newacro{SCM}{source code management}
\newacro{SDS}{spoken dialog system}
\newacro{SUS}{semantically unpredictable sentence}
\newacro{TTS}{text-to-speech synthesis}
\newacro{VER}{voicing error rate}
\newacro{WER}{word error rate}

\title{Creating New Language and Voice Components for the Updated MaryTTS Text-to-Speech Synthesis Platform}

\name{Ingmar Steiner, Sébastien Le Maguer}
\hypersetup{pdfauthor={Ingmar Steiner, Sébastien Le Maguer}}

\address{Saarland University \& DFKI GmbH\\
         \{steiner, slemaguer\}@coli.uni-saarland.de\\}

\abstract{%
  We present a new workflow to create components for the MaryTTS \acl{TTS} platform, which is popular with researchers and developers, extending it to support new languages and custom synthetic voices.
  This workflow replaces the previous toolkit with an efficient, flexible process that leverages modern build automation and cloud-hosted infrastructure.
  Moreover, it is compatible with the updated MaryTTS architecture, enabling new features and state-of-the-art paradigms such as synthesis based on \acp{DNN}.
  Like MaryTTS itself, the new tools are \ac{FOSS}, and promote the use of open data.
  \\\newline
}

\begin{document}
\setcounter{page}{3171}

\maketitleabstract
\acresetall

\section{Introduction}
\label{sec:introduction}

Over the last 15 years, MaryTTS \cite{schroeder2001} has become one of the reference systems for open source \ac{TTS}.
Today, it is actively used by researchers working in speech science, \ac{HCI}, and related fields, as well as by professional and enthusiast software developers in \ac{FOSS} or enterprise settings.
Its popularity is due in part to the number of languages and voices which are freely available as open resources, as well as the possibility of extending it to support new languages and building custom synthetic voices, or even integrating MaryTTS as a component into more complex applications, such as \ac{TTS} web services, accessibility software, or \acp{SDS}.
Because of its implementation in the Java programming language, MaryTTS can be used on any device or computer with a \ac{JRE}, and its modular design allows developers and users alike to inspect and customize the entire processing pipeline from input text to speech output.

However, the number of people who have participated in, and contributed to, MaryTTS development over the years has led to a complex and overburdened system.
Consequently, a reboot of the system became unavoidable;
until now, we focused on restructuring the system core and explained the philosophy behind the new architecture \cite{LeMaguer2017BC,LeMaguer2017ESSV}.

Independently, the process of creating new synthetic voices and support for new languages in MaryTTS has also fundamentally evolved since it was presented by \newcite{Pammi2010LREC}.
Therefore, the current paper presents the new language and voice building workflow for MaryTTS.

The remainder of the paper is structured as follows.
\Cref{sec:background} provides a brief background on build automation in MaryTTS.
In \cref{sec:new-language-support}, we present the new workflow to add support for a new language.
Then, in \cref{sec:voice-building}, we focus on the new voice building pipeline.
Finally, in \cref{sec:glob-proj-manag}, we present the reorganized source code and project hosting, particularly from a user perspective.

\section{Background}
\label{sec:background}


Development on MaryTTS has adopted several significant paradigms which had become best practice in Java-based software engineering in the years since the project's inception.
These include,
\begin{description}[nosep, wide, leftmargin=1em, font=\it, labelsep=0pt]
  \item[dependency management], where required software libraries are downloaded from cloud-based repositories,%
    \footnote{Examples of such dependencies in MaryTTS include third-party libraries for text tokenization (JTok), number expansion (ICU4J), and \ac{POS} tagging (OpenNLP).}
  \item[software testing], and
  \item[convention over configuration], where common standards are integrated into the software build lifecycle without the need for redundant specification.
\end{description}
In the latest version of MaryTTS, all of these aspects are managed through the \emph{Gradle} build automation tool.%
\footnote{\url{https://gradle.org/}}

The increase in flexibility and efficiency provided by Gradle is not limited to the development \enquote{under the hood}.
Rather, we leverage Gradle as a user-facing tool which replaces the custom applications previously required to add new languages to MaryTTS, or build new synthetic voices.
This shift removes numerous limitations on performance and functionality, and solves common, recurring problems with installing third-party tools and writing boilerplate code for new MaryTTS components.
At the same time, the text and speech data itself --- required to build new components --- can be managed as dependencies, and the components can be built, tested, and distributed more efficiently.

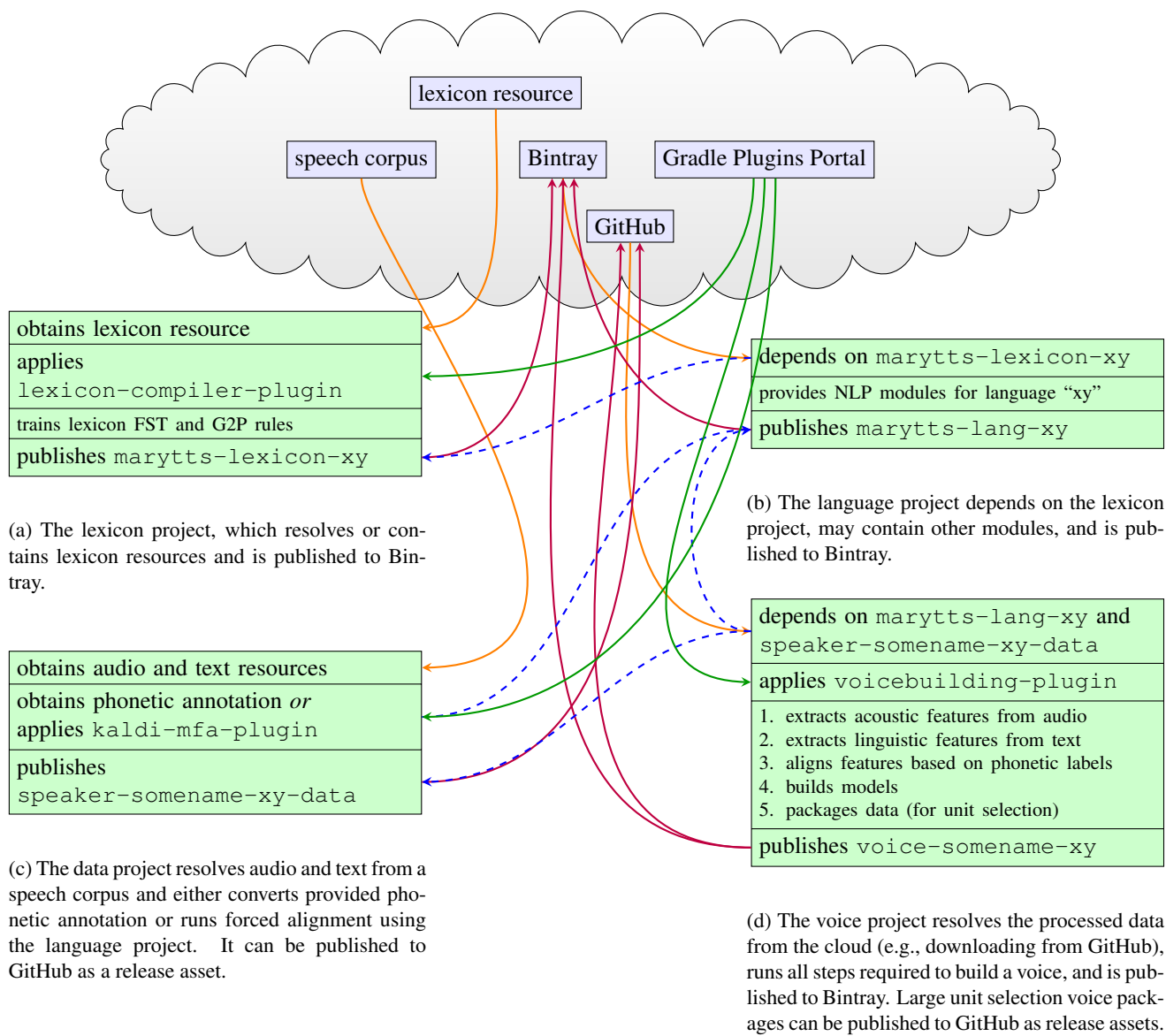
\begin{figure*}
  \begin{subfigure}{\linewidth}
    \centering
    \input{fig/cloud}
    \label{fig:overall-paragraph}
  \end{subfigure}
  \begin{subfigure}{0.75\columnwidth}
    \input{fig/lexicon-project}
    \caption{The lexicon project, which resolves or contains lexicon resources and is published to Bintray.}
    \label{fig:lexicon-project}
  \end{subfigure}
  \hfill
  \begin{subfigure}{0.75\columnwidth}
    \hfill
    \input{fig/language-project}
    \caption{The language project depends on the lexicon project, may contain other modules, and is published to Bintray.}
    \label{fig:language-project}
  \end{subfigure}
  \par
  \begin{subfigure}{0.75\columnwidth}
    \input{fig/data-project}
    \caption{The data project resolves audio and text from a speech corpus and either converts provided phonetic annotation or runs forced alignment using the language project.
      It can be published to GitHub as a release asset.}
    \label{fig:data-project}
  \end{subfigure}
  \hfill
  \begin{subfigure}{0.75\columnwidth}
    \hfill
    \input{fig/voice-project}
    \caption{The voice project resolves the processed data from the cloud (e.g., downloading from GitHub), runs all steps required to build a voice, and is published to Bintray.
      Large unit selection voice packages can be published to GitHub as release assets.}
    \label{fig:voice-project}
  \end{subfigure}
  \par
  \input{fig/pipeline-connectors}
  \caption{Overview of the complete workflow for a new language \enquote{xy} and synthetic voice components.
    Dashed blue arrows visualize the dependency of the voice project (\cref{fig:voice-project}) on the language and data projects (\cref{fig:language-project,fig:data-project}, respectively), the dependency of the data project on the language project, and the language project on the lexicon project (\cref{fig:lexicon-project}).
    All of these depend on the core MaryTTS runtime libraries (not shown), which are resolved from Bintray.
    Orange and purple arrows show the actual dependency resolution from, and publishing to, cloud-hosted services, respectively.
    Green arrows show plugins resolved from the Gradle Plugins Portal.
    Note that all or part of the cloud-hosted infrastructure (shown inside the cloud) could also be replaced by internal, non-public repositories.}
  \label{fig:pipeline}
\end{figure*}

An overview of the entire workflow to create new language and synthetic voice components is shown in \cref{fig:pipeline}.
However, this workflow can be broken up into several independent steps, which are described in the following sections.

\section{New Language Support}
\label{sec:new-language-support}

The purpose of a language component in MaryTTS is to allow the system to extract linguistic features from orthographic text using \ac{NLP}.
This includes, at the very least, the sequence of phonemes, i.e., the pronunciation, but typically also other features related to phonology and used for the prediction of acoustic parameters, such as segment duration and \ac{F0}.
Pronunciation prediction in MaryTTS is handled by a language-specific \enquote{Phonemiser} module, which looks up each text token in a lexicon and returns the sequence of phonemes.
For any \ac{OOV} tokens, the module falls back to rules for \ac{G2P} prediction.

To add support for a new language to MaryTTS, the first step is to define the set of phonemes to be used, along with their standard phonological features, based on the \ac{IPA}.
The next step is to obtain (or create) a lexicon resource, ideally a text file, spreadsheet, etc., containing a list of words with their orthographic and corresponding phonetic transcription.

Finally, the lexicon is automatically compiled into a \ac{FST}-based representation, relying in part on the WEKA toolkit \cite{WEKA}.
In the past, this was done using the custom \emph{TranscriptionTool} GUI application \cite{Pammi2010LREC}, which however suffers from various usability and performance issues.
To improve this situation, we have developed a Gradle plugin%
\footnote{\href{https://github.com/marytts/gradle-marytts-lexicon-compiler-plugin}{\nolinkurl{https://github.com/marytts/gradle-marytts-lexicon-compiler-}}
\href{https://github.com/marytts/gradle-marytts-lexicon-compiler-plugin}{\nolinkurl{plugin}}}
to convert the lexicon into the format required by MaryTTS.
Furthermore, we are currently developing a more state-of-the-art \ac{G2P} approach based on TensorFlow \cite{abadi2016tensorflow}, comparable to that of, e.g., \newcite{VanEsch2016IS}.

It is possible to create further \ac{NLP} modules for the new language component, handling text normalization to expand acronyms, numbers, and so on, into pronounceable representations, \ac{POS} tagging, etc.
Alternatively, MaryTTS can just fall back to generic modules for such tasks.
All of these modules are then combined to build the new language component, which will be used to process input text, and represents a dependency of the synthetic voice building, and ultimately, full \ac{TTS} in the new language.

\section{Voice Building}
\label{sec:voice-building}

Building a new synthesis voice for MaryTTS consists of three distinct stages,
\begin{enumerate*}[(a)]
  \item data preparation,
  \item feature extraction, and
  \item model building,
\end{enumerate*}
which are described in the following subsections.
All three stages are handled efficiently using Gradle plugins,%
\footnote{\url{https://github.com/marytts/gradle-marytts-voicebuilding-plugin}}
which wrap third-party tools and can run tasks in parallel where appropriate, speeding up the voice building process significantly compared to the old toolkit.

\subsection{Data Preparation}
\label{sec:data-preparation}

When preparing the recording of speech data intended to create a new synthesis voice, it is common practice to create a prompt list which covers the phonetic (and possibly prosodic) inventory of the corresponding language, as well as the content of the voice's domain.
These prompts are then read out by the voice talent over one or more recording sessions, preferably in a studio environment.

The previous voice creation toolkit \cite{Pammi2010LREC} promoted the use of a custom Java-based recording tool named \emph{Redstart}, which is able to display a sequence of prompts on a computer screen and record the user reading them through the computer's microphone.
While MaryTTS Redstart remains fully functional, it may not be usable in every recording scenario.
For instance, in a professional recording studio, the voice talent is typically recorded using a \ac{DAW}, and any visual presentation of prompts may only be possible using a separate computer.
In other cases, the goal may be to record a more fluent performance (such as an audiobook), and a user experience that forces the voice talent to pause for each prompt would be too disruptive.

Regardless of which text prompts are selected, or how they are recorded, the outcome of this process is a set of text and audio files with corresponding contents.
However, before these files can be used to build a synthetic voice for MaryTTS, they have to be phonetically annotated.
This step requires determining the pronunciation of each text prompt, i.e., the sequence of phonetic units, and mapping them to the recorded audio's time domain;
the process is related to \ac{ASR}, except that the expected content is known, and the sequence of phonetic units can be forced to align with the audio;
this is known as \emph{forced alignment}.
In the past, the MaryTTS voice building tools relied on integrating third-party tools for this task, including \emph{Sphinx-4} \cite{Walker2004SMLI}, \emph{HTK} \cite{HTKBook}, or the \emph{FestVox} tool \emph{EHMM} \cite{Prahallad2006ICASSP};
however, MaryTTS users often report problems installing or running them, and errors are difficult to solve.
More recently, \emph{Kaldi} \cite{Povey2011ASRU} has emerged as a leading \ac{ASR} toolkit, and it has been integrated into the \emph{Montreal Forced Aligner} tool \cite{McAuliffe2017IS}.
This tool in turn has been integrated into the MaryTTS data preparation workflow in the form of a Gradle plugin.%
\footnote{\url{https://github.com/marytts/gradle-marytts-kaldi-mfa-plugin}}
The pronunciation can be predicted using MaryTTS and collected into a custom dictionary for Kaldi, then acoustic models are trained from the recorded data, and the phonetic unit boundaries are aligned and stored in the form of Praat TextGrids;
this process is fully automated and can take a few minutes or hours, depending on the amount of recorded data.

Previously, the forced alignment process was described as part of the voice building process in MaryTTS \cite{Pammi2010LREC}, but it can be more appropriately regarded as a prerequisite.
While it is still possible to use both the forced alignment and voice building plugins in the same Gradle project, a more efficient workflow is to build a data \emph{artifact}, which is then available as a dependency for the proper voice building process.
Therefore, this stage can be skipped if a corpus of speech data is already available with appropriate orthographic and phonetic annotations.

%
%

\subsection{Feature Extraction}
\label{sec:feature-extraction}

At the core of the voice building process, the recorded speech data is converted to a \emph{feature representation}.
It is this feature representation which allows the use of machine learning techniques to train models to predict prosody and/or vocoder parameters from text during the actual \ac{TTS} process in the runtime system.

The feature extraction stage of the voice building process yields a combination of frame-wise feature vectors from acoustic analysis of the audio, and time-aligned symbolic features based on linguistic analysis of the corresponding text;
the alignment is based on the phonetic annotation obtained in the data preparation (cf.\ \cref{sec:data-preparation}).

Acoustic features include \ac{F0}, tracked using Praat \cite{Praat}, and \acp{MFCC}, extracted using the \ac{EST} \cite{EST}.
The linguistic features are obtained depending on the MaryTTS language component for the corresponding language.
When creating a synthetic voice for a new language, this is where the new language component built previously  (cf.\ \cref{sec:new-language-support}) is used.
The linguistic features extracted and assigned to the feature vectors include several related to phonology (e.g., distinctive features, position in the syllable, stress, accent), syntax (e.g., \ac{POS}, distance to phrase and sentence boundaries), and --- optionally --- speaking style \cite{Steiner2010SSW,Charfuelan2013IS}, information density \cite{LeMaguer2016SP}, or other high-level context features.

\subsection{Model Building}
\label{sec:model-building}

Depending on the underlying synthesis paradigm, it is possible to build a \emph{unit selection} voice or a \emph{statistical parametric synthesis} voice.

\subsubsection{Unit Selection}
\label{sec:unit-selection}

Unit selection synthesis concatenates halfphone-sized snippets of natural speech selected from a voice database, given target features computed for an input utterance.
The output can sound very natural, but often suffers from audible glitches when synthesizing out-of-domain utterances, and prosody control is limited.
Moreover, the voice database can be very large, as it contains the actual audio data.

Building a unit selection voice for MaryTTS involves storing the feature representation and related metadata for each unit, training statistical models for sparse prosody prediction, and packaging these along with the actual audio data.
We have created a Gradle plugin which wraps some of the old toolkit's components to assemble unit selection voices which are backward-compatible with the current stable release of MaryTTS (v5.2).
In addition, we are developing new build tools to support audio compression and enable prosody modeling and target feature prediction using \acp{HMM} or \acp{DNN}, paving the way for state-of-the-art \enquote{hybrid} \ac{TTS}.

\subsubsection{Statistical Parametric Synthesis}
\label{sec:parametric-synthesis}

MaryTTS has supported statistical parametric synthesis for numerous years, using a Java port of the \ac{HTS} engine API%
\footnote{\url{http://hts-engine.sourceforge.net/}} with a \ac{MLSA} vocoder.
Although such synthesis can sound rather buzzy and unnatural, these \ac{HMM}-based voices offer higher flexibility and more consistent quality than unit-selection synthesis, as well as a much smaller memory footprint.
However, some drawbacks are
\begin{enumerate*}[(a)]
  \item that building \ac{HMM}-based voices for MaryTTS has a high technical overhead, and
  \item that the Java port has become quite outdated, while \ac{HTS} development has seen significant progress.
\end{enumerate*}
The former has been mitigated by providing a consistent, pre-configured \emph{Docker} container, while to address the latter, we are developing completely new functionality.
This includes the possibility to train models for third-party frameworks such as \emph{Merlin} \cite{Merlin} and to allow other vocoders to be used, including \emph{STRAIGHT} \cite{Kawahara1999} or \emph{WORLD} \cite{Morise2016TIS}.

The parametric voice building process comprises three stages: the input and output feature packing, the
model training, and the voice configuration generation.
The voice configuration generation is similar to the unit selection voice building part (cf.\ \cref{sec:unit-selection}).
The output feature packing goal is just to adapt the acoustic features (e.g., \ac{MGC}, \ac{F0}, \ac{BAP}, etc.) to be compatible with the process used to train the models.
Currently this means computing the delta and delta-delta coefficients and generating the binary observation vector for each utterance.
The input feature packing consists of calling MaryTTS with a serializer dedicated to the training process.

The model training is a specific plugin implementing the process to train the models needed for the synthesis stage.
We have developed a Gradle plugin dedicated to train HTS models (\ac{HMM}-\acs{GMM} or \ac{HMM}-\ac{DNN}).%
\footnote{\url{https://github.com/marytts/gradle-hts-voicebuilding-plugin}}
This plugin can be adapted to the kind of parametric synthesis model or system we want to use.

\subsection{New Configuration Mechanism}
\label{sec:new-conf-mech}

Previously a configuration was attached to an artifact to configure the different modules.
Moving forward, we consider three levels of configuration: the default configuration, the voice configuration, and the user configuration.
The first of these is given in the module itself.
The voice configuration corresponds to the parametrization of each module used during the voice building process and has priority over the default configuration.
Finally, a user configuration can be specified at runtime, to override the other configurations.

\section{Global Project Management}
\label{sec:glob-proj-manag}

Refactoring the core system and of the voice building process has allowed us to separate the \ac{SCM} for each language and each voice project.
Therefore, each language and voice can have its own \ac{SCM} repository hosted on GitHub,%
\footnote{e.g., \url{https://github.com/marytts/voice-dfki-spike}}
while the released artifacts are published to Bintray%
\footnote{\url{https://bintray.com/marytts/marytts}}
and indexed in JCenter.%
\footnote{\url{https://bintray.com/bintray/jcenter}}
Any large data objects (specifically unit selection audio data) can be hosted on GitHub as release assets.

This makes the custom \emph{Component Installer} GUI from previous MaryTTS versions obsolete, and allows us to replace it with a lightweight wrapper around the dependency management.
A user can install and run MaryTTS voices and language components simply by executing Gradle tasks with the corresponding names;
this is demonstrated by a new web installer for MaryTTS.%
\footnote{\url{https://github.com/marytts/marytts-installer}}

Meanwhile, developers and researchers looking to integrate MaryTTS into their projects, only need to declare a dependency on the desired voice artifacts, and this will automatically resolve all transitive dependencies on the corresponding languages and other libraries.

\section{Conclusion}
\label{sec:concl-persp}

In conclusion, we have presented a new language and voice building workflow designed for the updated MaryTTS system.
We have detailed our reliance on the Gradle build automation tool, which provides a much more efficient and powerful framework via its extensible plugin system than the previous toolkit.
We have also seen that the language components maintain the same concepts as in previous versions, but the methodologies used are updated.
Finally, we have described the redesigned and extended voice building process, as well as our leverage of cloud-based infrastructure for hosting and distribution.

The next stage is to integrate the new MaryTTS core, state-of-the-art synthesis paradigms, and the new build system more deeply to provide the fully modular, modern \ac{TTS} platform we are aiming for.
Moreover, we are working to release the first preview of MaryTTS v6.0 in the coming months.

\section{Acknowledgements}

This work was funded by the German Research Foundation (DFG) under grants EXC~284 and SFB~1102.

\section{Bibliographical References}
\label{main:ref}

\bibliographystyle{lrec}
\bibliography{biblio}

\listoftodos

\end{document}

%% file: fig/cloud.tex
\begin{tikzpicture}[%
  remember picture,
  cloudthingie/.style={%
    draw,
    fill=cloudthingiecolor
  }
]

\node (lexicon-resource) [cloudthingie] at (-6,1) {lexicon resource};
\node (speech-resource) [cloudthingie] at (-8,0) {speech corpus};
\node (bintray) [cloudthingie] at (-5,0) {Bintray};
\node (gradle-plugins) [cloudthingie] at (-2,0) {Gradle Plugins Portal};
\node (github) [cloudthingie] at (-4,-1) {GitHub};

\scoped[on background layer]
\node (cloud) [%
  draw,
  top color=white,
  bottom color=gray!20,
  cloud,
  cloud puffs=36,
  cloud puff arc=150,
  aspect=4,
  inner sep=0,
  fit=(lexicon-resource) (speech-resource) (bintray) (gradle-plugins) (github)
] {};
\end{tikzpicture}

%% file: fig/lexicon-project.tex
\begin{tikzpicture}[remember picture]
  \node (lexicon-project) [%
      draw,
      rectangle split,
      rectangle split parts=4,
      rectangle split part fill=projectcolor,
      text width=0.95\columnwidth
  ] {%
    obtains lexicon resource
    \nodepart{two}
    applies \texttt{lexicon-compiler-plugin}
    \nodepart{three}
    \smaller
    trains lexicon \acs{FST} and \acs{G2P} rules
    \nodepart{four}
    publishes \texttt{marytts-lexicon-xy}
  };
\end{tikzpicture}

%% file: fig/language-project.tex
\begin{tikzpicture}[remember picture]
  \node (language-project) [%
      draw,
      rectangle split,
      rectangle split parts=3,
      rectangle split part fill=projectcolor,
      text width=0.95\columnwidth
  ] {%
    depends on \texttt{marytts-lexicon-xy}
    \nodepart{two}
    \smaller
    provides \acs{NLP} modules for language \enquote{xy}
    \nodepart{three}
    publishes \texttt{marytts-lang-xy}
  };
\end{tikzpicture}

%% file: fig/data-project.tex
\begin{tikzpicture}[remember picture]
  \node (data-project) [%
      draw,
      rectangle split,
      rectangle split parts=3,
      rectangle split part fill=projectcolor,
      text width=0.95\columnwidth
  ] {%
    obtains audio and text resources
    \nodepart{two}
    obtains phonetic annotation \emph{or} \\
    applies \texttt{kaldi-mfa-plugin}
    \nodepart{three}
    publishes \texttt{speaker-somename-xy-data}
  };
\end{tikzpicture}

%% file: fig/voice-project.tex
\begin{tikzpicture}[remember picture]
  \node (voice-project) [%
      draw,
      rectangle split,
      rectangle split parts=4,
      rectangle split part fill=projectcolor,
      text width=0.95\columnwidth
  ] {%
    depends on \texttt{marytts-lang-xy} and \texttt{speaker-somename-xy-data}
    \nodepart{two}
    applies \texttt{voicebuilding-plugin}
    \nodepart{three}
    \parbox{\linewidth}{%
      \smaller
      \begin{enumerate}[nosep, wide]
        \item extracts acoustic features from audio
        \item extracts linguistic features from text
        \item aligns features based on phonetic labels
        \item builds models
        \item packages data (for unit selection)
      \end{enumerate}
    }
    \nodepart{four}
    publishes \texttt{voice-somename-xy}
  };
\end{tikzpicture}

%% file: fig/pipeline-connectors.tex
\begin{tikzpicture}[remember picture, overlay, ->, >=stealth, thick]

  \begin{scope}[color=orange]
    \draw [out=270, in=0] (lexicon-resource.south) to (lexicon-project.text east);
    \draw [out=270, in=180] (bintray.south) to (language-project.text west);
    \draw [out=270, in=0] (speech-resource)
      .. controls ($(speech-resource.south) + (0,-1)$)
      and ($(data-project.text east) + (4,0)$)
      .. (data-project.text east);
    \draw [out=270, in=180] (github.south) to (voice-project.text west);
  \end{scope}

  \begin{scope}[color=purple]
    \draw [out=0, in=270] (lexicon-project.four east) to (bintray.240);
    \draw [out=180, in=270] (language-project.three west) to (bintray.300);
    \draw [out=0, in=270] (data-project.three east) to (github.300);
    \draw [out=180, in=270] (voice-project.four west) to (bintray.south);
    \draw [out=180, in=270] (voice-project.four west) to (github.240);
  \end{scope}

  \begin{scope}[color=blue, dashed]
    \draw [out=180, in=0] (language-project.text west) to (lexicon-project.four east);
    \draw [out=0, in=180] (data-project.two east) to (language-project.three west);
    \draw [out=180, in=180] (voice-project.text west) to (language-project.three west);
    \draw [out=180, in=0] (voice-project.text west) to (data-project.three east);
  \end{scope}

  \begin{scope}[color=green!60!black]
    \draw [out=270, in=0] (gradle-plugins.240) to (lexicon-project.two east);
    \draw [out=270, in=0] (gradle-plugins.300) to (data-project.two east);
    \draw [out=270, in=180] (gradle-plugins.south) to (voice-project.two west);
  \end{scope}
\end{tikzpicture}